\let\origvec\vec
\documentclass[runningheads]{llncs}

\let\vec\origvec
\usepackage{comment}
\usepackage{url}
\usepackage{times}
\usepackage{epsfig}
\usepackage{graphicx}
\usepackage{amsmath,amssymb}
\usepackage{color}
\usepackage{multirow}
\usepackage{booktabs}
\begin{document}
\pagestyle{headings}
\mainmatter
\def\ECCVSubNumber{****}

\title{A Zero-Shot based Fingerprint Presentation Attack Detection System}
\titlerunning{ZSPAD}
\authorrunning{Liu et al.}
\author{Haozhe Liu\inst{1}\thanks{Haozhe Liu and Wentian Zhang contribute the same to this article} \and Wentian Zhang\inst{1}$^\star$ \and Guojie Liu\inst{1} \and Feng Liu\thanks{The correspondence author is Feng Liu and the email: feng.liu@szu.edu.cn}\inst{1}}
\institute{$^1$The National Engineering Laboratory for Big Data System Computing Technology; \\
The Guangdong Key Laboratory of IntelligentInformation Processing; \\
College of Computer Science and Software Engineering, \\
Shenzhen University, Shenzhen 518060, China.}
\maketitle

\begin{abstract}
With the development of presentation attacks, Automated Fingerprint Recognition Systems(AFRSs) are vulnerable to presentation attack.
Thus, numerous methods of presentation attack detection(PAD) have been proposed to ensure the normal utilization of AFRS.
However, the demand of large-scale presentation attack images and the low-level generalization ability always astrict existing PAD methods' actual performances.
Therefore, we propose a novel Zero-Shot Presentation Attack Detection Model to guarantee the generalization of the PAD model.
The proposed ZSPAD-Model based on generative model does not utilize any negative samples in the process of establishment, which ensures the robustness for various types or materials based presentation attack.
Different from other auto-encoder based model, the Fine-grained Map architecture is proposed to refine the reconstruction error of the auto-encoder networks and a task-specific gaussian model is utilized to improve the quality of clustering.
Meanwhile, in order to improve the performance of the proposed model, 9 confidence scores are discussed in this article.
Experimental results showed that the ZSPAD-Model is the state of the art for ZSPAD, and the MS-Score is the best confidence score.
Compared with existing methods, the proposed ZSPAD-Model performs better than the feature-based method and under the multi-shot setting, the proposed method overperforms the learning based method with little training data. When large training data is available, their results are similar.
\end{abstract}
\section{Introduction}
\label{sec:Intro}
The rise in the number of intelligent device terminals coupled with increased convenient payment tools, has resulted in an exponential growth in Automated Fingerprint Recognition Systems(AFRSs) \cite{cappelli2010minutia,chugh2019oct,yin20193d}.
The purpose of AFRSs is to ensure a reliable and accurate user authentication \cite{chugh2019oct} for various applications such as mobile payments \cite{das2018recent} and access control \cite{hutz2019digital}.
However, current AFRSs are easily spoofed by presentation attack (PA) \cite{goicoechea2016evaluation} made from silica gel or other low cost materials \cite{liu2019high,sousedik2014presentation}.

For presentation attack detection(PAD), numerous methods have been proposed and they can be roughly divided into two categories.
The first category of method is on the basis of software. These methods exploit dynamic behaviors of live fingertips(e.g., perspiration \cite{jia2007new} and ridge distortion \cite{antonelli2006fake}) and the static features (e.g. textural features) \cite{lee2009fake,nikam2008fingerprint} for PAD.
For example, the approach proposed by Antonelli et al. \cite{antonelli2006fake} required users to move the finger while pressing it against the scanner surface and detects the PA based on skin distortion. Nikam et al. \cite{nikam2008fingerprint} proposed the textural-characteristic-based method, whose classification rate ranges from ~94.35\% to ~98.12\%.
Although it is convenient to deploy these software based approaches and the correspinding cost is at a low level,
the research \cite{bontrager2018deepmasterprints} shows that these software based methods can be easily spoofed by the Generative Adversarial Networks(GAN) \cite{bontrager2018deepmasterprints}.
\begin{figure}
  \centering
  \includegraphics[width=.88\textwidth]{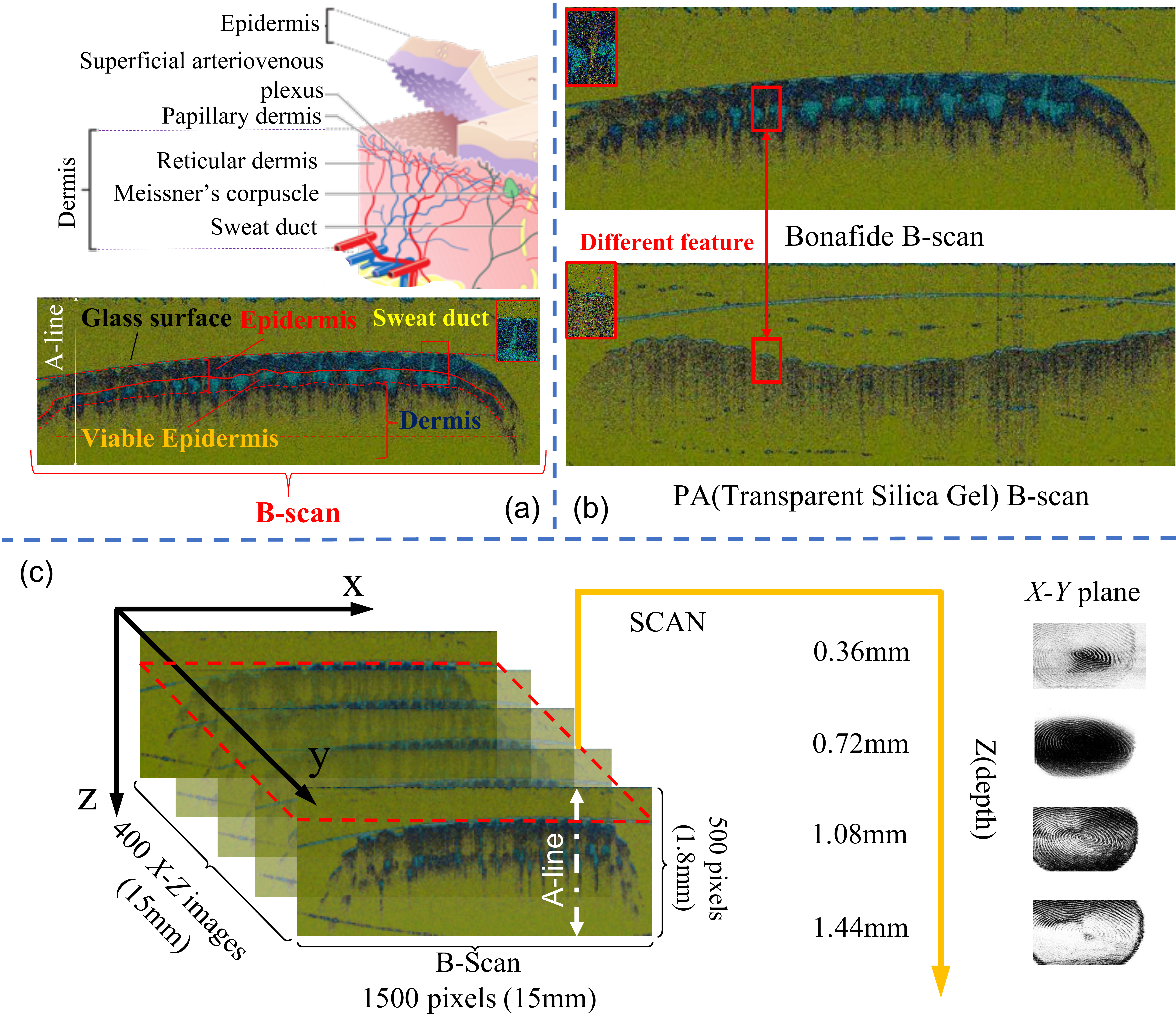}
  \caption{An example of cross-section fingerprint image of Bonafide and PA:
           (a). The cross-section fingertip image marked with biological structure \cite{skinlayer}.
           (b). The cross-section images of Bonafide (in the first row) and Presentation Attack (in the second row) made by transparent silica gel.
           (c). The X-Y tomographic images of the internal fingerprint \cite{liu20193d}.
           }
  \label{fig:sample}
\end{figure}

Different from the software based methods, hardware based methods pay more attention to capture the liveness characteristics from fingertips by special sensors (e.g., multispectral sensor \cite{nixon2004novel,reddy2008new}, high-resolution sensor \cite{zhao2010adaptive}, binocular stereo vision sensor \cite{liu2015study} and Optical Coherence Technology (OCT) \cite{huang1991optical}).
Nixon et al.\cite{nixon2004novel} proposed a spectroscopy-based device to obtain the spectral feature of the fingertip, which is proved to be able to distinguish between the PA and real fingerprint.
The pulse-oximetry-based device, which involves the source of light originating from a probe at two wavelengths, was proposed by Reddy et al. \cite{reddy2008new} to capture the percentage of oxygen in the blood as well as the heart pulse rate for liveness detection (i.e. PAD).
Zhao et al. \cite{zhao2010adaptive} proposed a method for the pore extraction using high-resolution sensor and applied it for liveness detection.
The 3D fingerprint using binocular stereo vision proposed by Liu and Zhang can be utilized for liveness detection.
Liu et al. \cite{liu2019high} proposed a method to obtain the depth-double-peak feature and sub-single-peak feature for PAD using optical-coherence-technology-based device, which achieves 100\% accuracy over all four types of artificial fingerprints.
Among them, OCT-based device extracts the internal surface of fingerprint and it can be utilized for liveness detection and identification at the same time.

The difference between OCT and other devices has led to the emergence of PAD using OCT as an active area of research \cite{bossen2010internal,chang2008optical,cheng2006artificial,chugh2019oct,dubey2007fingerprint,huang1991optical,liu2019high,marasco2015survey}.
OCT-based device provides an internal representation of the fingertip skin rather than a simple feature (e.g., spectral feature, the percentage of oxygen and heart pulse rate).
And the depth information and representation formation of such internal fingerprints has excellent capability of PAD \cite{chugh2019oct,liu2019high}.
Figure. \ref{fig:sample} shows the samples captured by OCT. The biological fingertip structure of bonafide(real data) including Epidermis and Dermis shown in Figure. \ref{fig:sample}(a) are captured by OCT in the form of B-scan and A-line.
Each A-line, which reflects the capability of sensor to image the subsurface characteristics of the skin, reaches to a penetration depth and corresponds to pixels along depth direction.
B-scan, namely a cross-section image, is then formed by all A-lines scanning along the horizontal direction.
For 3D scanning, a slow scanning galvanometer is employed to obtain 400 B-scans.
Thus, the 3D fingerprint image captured by our device consists of four hundreds longitudinal(X-Z) fingertip images with spatial size of 1500 $\times$ 400 $\times$ 500 pixels to quantify a real fingerprint area of 15mm $\times$ 15mm $\times$ 1.8mm, as shown in Fig. \ref{fig:sample}(c).
As illustrated in Figure. \ref{fig:sample}(b), the presentations of bonafide and PA are different in B-scan.
Therefore, this representation is generally useful for PAD.
\begin{figure}
  \centering
  \includegraphics[width=.86\textwidth]{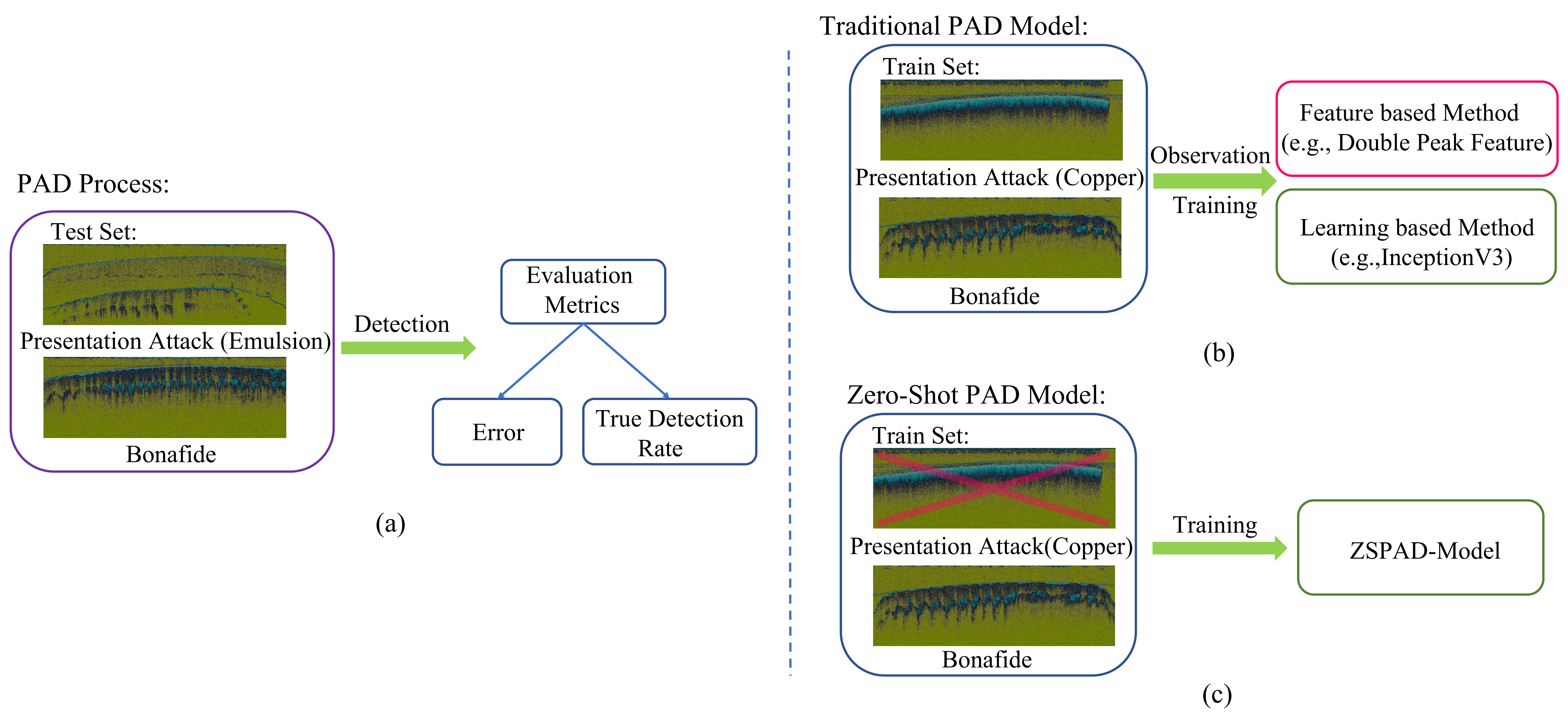}
  \caption{The pipelines of PAD:
           (a). The process of PAD.
           (b). Traditional PAD Model.
           (c). Zero-Shot based PAD Model.
           }
  \label{fig:zspad}
\end{figure}

For the topic of PAD, there are two challenges:
the first challenge here is the feature extraction from the B-scans to distinguish the PAs from bonafides. The second challenge is the establishment of the auto-PAD model, which maps the features to the liveness confidence.
Ideally, for better generalization, a model for PAD using OCT must learn to discover the alignments between the PA and the bonafide.
For example, in Fig. \ref{fig:zspad}(a)(b), the model is determined by the training set including PA made by copper(seen) as well as bonafide and we want the model to distinguish as well PA made by emulsion(unseen) \& bonafide.
However, current evaluation methodology \cite{chugh2019oct,liu2019high} focuses only on class-based PAD, i.e., seen presentation attacks made of various instruments and bonafides are classified into different categories by the model. However, unseen presentation attack is difficult to detect. For example, in Fig. \ref{fig:zspad}(a), unseen presentation attack made of emulsion has different characteristics from bonafide and PA made of copper and there is no guarantee that the model can detect emulsion PA. This leads to the dependence of the model on data and the security risks of AFRSs.

Considering the PAD problem, presentation attack instruments (PAIs) are difficult to collect and it is impossible to master all potential PAIs.
Therefore, the model with low data dependence is more suitable for PAD.
In this paper, we propose a Zero-Shot framework for PAD (ZSPAD) using OCT.
Since we can get a large number of bonafide data in the AFRSs, the proposed ZSPAD only requires bonafide samples to complete PAD. Without any PA for reference, the method can have better generalization and solve the detection problem caused by the gap between different PAIs (presentation attack instruments).

The paper is organized as follows: In Section \ref{sec:RW}, we give a brief overview of the state-of-the-art techniques in PAD using OCT and Zero-Shot learning(ZSL). Subsequently, in Section \ref{sec:AEMZSPAD}, we introduces the general idea for ZSPAD, the corresponding specific adaptations to our model and the feasible confidence scores for presentation attack detection. Then Secition \ref{sec:ERA} shows the empirical evaluation of our proposed method in ZSPAD setting and the pros \& cons of existing state-of-the-art PAD methods. Finally, we summarize the topic of ZSPAD as well as our proposed approach in Section \ref{sec:Con}.
\section{Related Work}
\label{sec:RW}
Since we propose a Zero-Shot framework for the PAD task, we briefly review the literature from both PAD using OCT as well as Zero Shot learning in this section.

As shown in Fig. \ref{fig:zspad}(b), there are two typical approaches \cite{liu2019high,chugh2019oct} for presentation attack detection using OCT \cite{bossen2010internal,chang2008optical,cheng2006artificial,chugh2019oct,dubey2007fingerprint,huang1991optical,liu2019high,marasco2015survey}.
The first one is based on feature extraction. Liu et al. \cite{liu2019high} defined two features for PAD, namely depth-double-peak and sub-single-peak, by the observation and comparison of the A-line between PA and bonafide. Depth-double-peak feature refers that there must be two and only two peaks in one A-line which reflects the pysical structure of real fingertip. While sub-single-peak feature means that there should be a peak in the A-line intercepted before the maximum peak.
This method achieves 100\% accuracy over all four types of PAIs.
The second category of PAD is based on learning model. Chugh and Jain \cite{chugh2019oct} develop and evaluate a PA detector based on deep convolutional neural network(CNN). Input to CNN are local patches extracted from the B-scans captured from THORLabs Telesto series spectral-domain OCT scanner. Chugh and Jain's approach achieves a true detection rate(TDR) of 99.73\% on a database of 3,413 bonafides(8 different PAIs) and 357 PA OCT scans. Among them, the feature based method is sensitive to the noise and the its robustness is poor.
The learning based method are data-dependent, which limits the scope of the application.
In the topic of PAD, it is hard to collect presentation attack instruments(PAIs). Therefore, the robustness and generalization of these methods can not be guaranteed.

It is promising to propose a PAD method without the dependence on any PA data.
This classification task without class samples belongs to the category of Zero-Shot learning.
Zero-Shot learning has been applied in various research areas
\cite{akata2015evaluation,changpinyo2016synthesized,Felix_2018_ECCV,kumar2009attribute,Li_2019_ICCV,Wang_2019_ICCV,Song_2018_CVPR},
such as face verification \cite{kumar2009attribute}, object categorization \cite{Bansal_2018_ECCV} and image retrieval \cite{Yelamarthi_2018_ECCV}.
The tasks of detecting the classes without any seen data are called Zero-Shot learning (ZSL).
Therefore, the main challenge of ZSL is how to generalize the models to detect the novel object categories without accessing any labeled instances of these categories \cite{fu2018recent}.
ZSL generally includes inductive ZSL \cite{akata2015evaluation,changpinyo2016synthesized} and transductive ZSL \cite{fu2015transductive,yu2017transductive}. For the inductive ZSL, only data of the source categories are available during the training phase.
While the transductive ZSL method is under an open-set setting that both the labeled source images and unlabeled target images are available for training \cite{yu2017transductive}.
Unfortunately, it is difficult for both inductive and transductive methods to be directly applied for the unseen class detection. They should learn domain-invariance and generalize feature representation for domain adaptation \cite{fu2018recent}. The feature representation for ZSL is usually based on semantic attributes including parts, shape, materials, etc \cite{fu2015transductive,kumar2009attribute,Yelamarthi_2018_ECCV}. For example, an attribute 'has wings' refers to the intrinsic characteristic of an instance or a class (e.g., bird) \cite{fu2015transductive}.

However, for presentation attack detection, PA belongs to unseen data. Therefore, it is very difficult for PAD to find the reasonable semantic attributes to distinguish between bonafides and PAs.
As shown in Fig. \ref{fig:zspad}(c), we propose a novel Zero-Shot framework under a Zero-PA setting, where the Zero-Shot PAD model is trained without any PAs.
Specifically, an auto-encoder(AE) network is trained to reconstruct bonafide B-scans, and the trained model is utilized to evaluate the reconstruction error of PA and bonafide B-scan.
Since there is high-level noise and invalid areas in B-scan image disturbing the reconstruction error, we propose a Fine-grained Map architecture(FineMap) to provide a constraint for B-scan in order to obtain small reconstruction error.
Considering that each fingertip has multiple B-scans, we utilized the Gaussian Model to estimate the probability density of the reconstruction errors of a fingertips. Finally, three detection methods including, statistical feature, probability density and divergence based confidence scores are proposed for ZSPAD.
Our experimental results show that the proposed ZSPAD-Model performs better than the feature-based method. And under the multi-shot setting, the proposed method overperforms the learning based method with little training data. When large training data is available, their results is similar.

\section{Presentation Attack Detection using ZSPAD-Model}
\label{sec:AEMZSPAD}
\begin{figure}[!htbp]
  \centering
  \includegraphics[width=.86\textwidth]{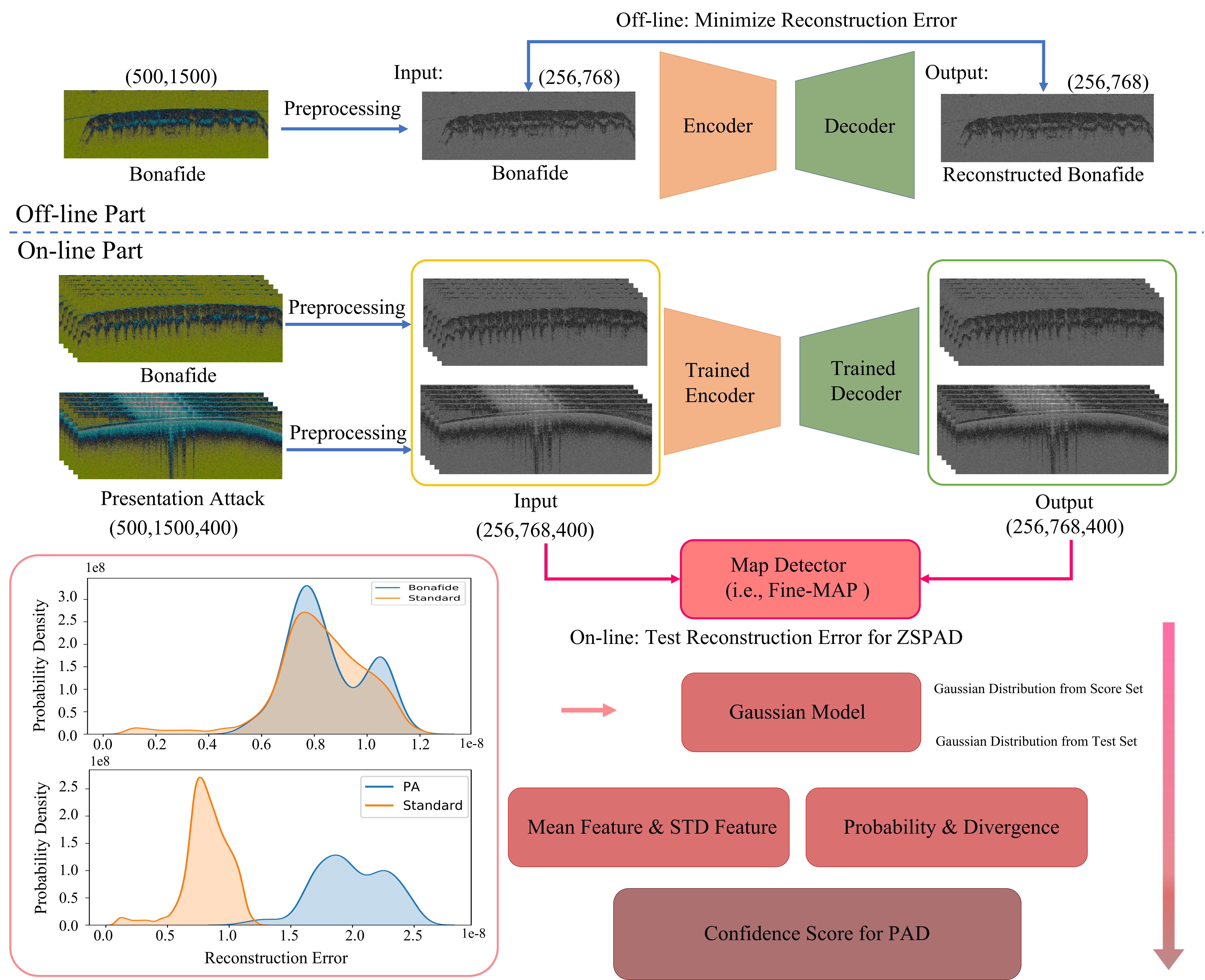}
  \caption{An overview of the proposed Zero-Shot based fingerprint presentation attack detection approach utilizing the auto-encoder network.}
  \label{fig:generalpipeline}
\end{figure}

As shown in Fig.\ref{fig:zspad}(a)(b), the existing approaches mentioned in the Section \ref{sec:Intro} and Section \ref{sec:RW} for presentation attack detection requires PAIs to establish the feature extraction or supervised neural network model.
Therefore, the existing approaches can not generalize well to the ZSPAD task.
We now propose an Zero-Shot learning based method for ZSPAD.
Fig. \ref{fig:generalpipeline} shows the pipeline of our proposed method, namely ZSPAD-Model, for this problem.
ZSPAD is divided into two parts, off-line part and on-line part. The off-line part is utilized to train the auto-encoder network, both input and output are the B-scans of bonafides.
The on-line part is used to test the samples.
The main motivation for such an approach is that we can get the precise probability distribution of the bonafide B-scans through the auto-encoder network, and the probability distribution of the bonafide can be utilized for the detection of PA.
Due to the interference of the invalid area and the speckle noise of the B-scans, it is difficult for simple auto-encoder structure to estimate accurate probability distribution.
Additionally, a series of B-scans are derived from OCT \cite{liu2019high}, when the fingertips or PAIs are scanned.
But the auto-encoder network can not utilize the multi B-scans' information in a reasonable way. Therefore, we propose a series of architectures and approaches to solve the above issues.

Below, we first introduce the off-line training part for ZSPAD in Section \ref{sec:AEN}.
Specifically, we provide a preprocessing method to reduce the speckle noise.
An auto-encoder network is proposed to evaluate the probability distribution.
In Section \ref{sec:ROI}, we present a PAD-specific approaches to promote the precision of the gaussian model using Fine-grained Map.
Finally, we describe the confidence scores based on the statistical features and probability distribution for PAD using Gaussian Model in Section \ref{sec:CSPAD}.
\subsection{Off-line Part for ZSPAD}
\label{sec:AEN}
In this section, we first introduce the on-liine part for PAD as shown in Fig. \ref{fig:generalpipeline}. The proposed off-line part of PAD approach includes two stages: a preprocessing stage(image resize and noise removal) and an auto-encoder network(estimate the probability distribution of the single B-scan).
\subsubsection{Preprocessing}
Optical Coherence Tomography(OCT) scans(B-scans) are RGB images with height $= 500$ pixels and width $=1500$ pixels(see Fig.\ref{fig:generalpipeline}).
These images contain speckle noise, which reduces the performance subsequent processing. Firstly, in order to reduce the computation and memory for subsequent processing, we utilize the bilinear interpolation to resize the B-scan to a smaller size(height$= 256$,width $=768$) and convert them to gray images.
Then, inspired by \cite{chugh2019oct}, we employ Non-Local Means denoising \cite{buades2011non} to remove noise by replacing the intensity of a pixel with an average intensity of similar pixels that may not be present close to each other(non-local) in the image.
More concretely, the implementation of Non-Local Means denosing is based on opencv with $filter$-$Strength = 19$, $templateWindowSize=7$ and $searchWindowSize=21$.
\subsubsection{Auto-encoder Network for ZSPAD}
A general idea to achieve ZSPAD is to find the probability distribution of bonafide $\mathbb{P}_b$ and images, which do not fit $\mathbb{P}_b$, are identified as presentation attack. Therfore, ZSPAD problem can be described as finding a model $F$, whose input is $x$ and output is the $\mathbb{P}_b(x)$,
\begin{align}
  F(x) = \mathbb{P}_b(x)
\end{align}
where $F$ is a generative model. Several models such as Gaussian Mixture Model(GMM) and Hidden Markov Model(HMM) can be utilized for this problem. However, considering the size of the B-scans and the complexity of the ZSPAD, these methods can not handle the high-dimensional data $x$ (B-scan), and are difficult to get the precise $\mathbb{P}_b(x)$.
Thus, we propose an auto-encoder based Model $F_{AE}$. Fig.\ref{fig:generalpipeline} shows the structure of auto-encoder network. The auto-encoder network is composed of two parts, encoder and decoder. Encoder encodes the input $x$ and extracts the features of the $x$.
Decoder utilizes these features to reconstruct the input $x$. Both encoder and decoder consists of multi convolution layers. Fig. \ref{fig:pipeline} shows the architecture of the auto-encoder network utilized in this paper.
Specifically, decoder consists of 5 Resnet-blocks \cite{he2016deep} and each block contains three atrous convolution layers(rate=1,2,5, kernel size=(3,3)) and one convolution layer (kernel size=(3,3), step=2).
Encoder is composed of 6 block and each block consists of two layers, bilinear resize layer and convolution layer (kernel size=(3,3)). The target of $F_{AE}$ is given by,
\begin{align}
  F_{AE}(x,\theta) = x
\end{align}
where the both input and output of $F_{AE}$ is $x$ and the $\theta$ is the parameters of auto-encoder network.
This network's update strategy is described as,
\begin{align}
  \underset{\theta}{min} \quad R_{AE}(x,\theta) = ||F_{AE}(x,\theta) - x||_2
\end{align}
where $R_{AE}$ is the reconstruction error for the $F_{AE}$.
As can be seen, $F_{AE}$ does not provide the probability distribution $\mathbb{P}$ of $x$ explicitly. But $F_{AE}$ has a such property that, the probability distribution of $x$ is related to reconstruction error $R_{AE}$. Given an unseen B-scan $x_i$, when the reconstruction error of $x_i$ is small, $x_i$ is close to the $\mathbb{P}$, when the reconstruction error is large, $x_i$ is far away from $\mathbb{P}$. Thus, the ZSPAD problem is further transformed into making the probability $\mathbb{P}$ of auto-encoder network close to $\mathbb{P}_b$.
To solve this problem, only the bonafide data set $ S_{model} $ is utilized to train the $F_{AE}$ and it is described as,
\begin{align}
  \underset{\theta}{min} \quad  R_{AE}(x,\theta) \quad s.t. \quad x \in S_{model}
\end{align}
Then, we can get the reconstruction error $R^*_{AE}(x)$ of the trained auto-encoder network.
Intuitively, when $R^*_{AE}(x)$ is larger, the probability of $x$ being PA is higher, and when $R^*_{AE}(x)$ is smaller, the probability of $x$ being bonafide is higher.
However, as shown in Fig. \ref{fig:sample}, there is invalid area and speckle noise in the B-scans, which disturbs the precision of $R^*_{AE}(x)$. Hence, we propose a Fine-grained Map architecture to refine the reconstruction error, which is presented in the next section.
\subsection{Refine ZSPAD Result using FineMap Architecture}
\label{sec:ROI}
In order to reduce the reconstruction error provided by trained auto-encoder network and the influence from invalid area of B-scans, we proposed Fine-grained Map (FineMap) architecture in this section.
\begin{figure}
  \centering
  \includegraphics[width=.9\textwidth]{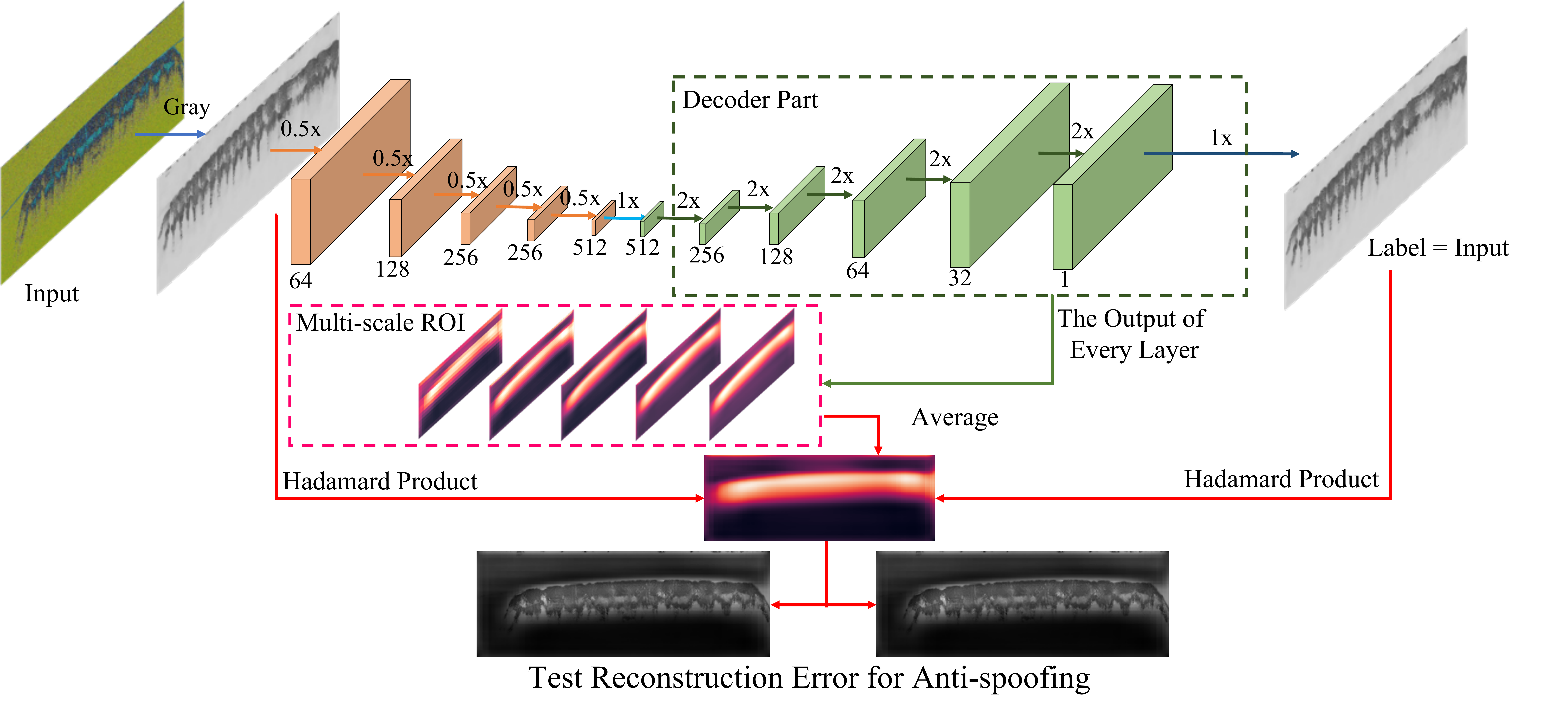}
  \caption{The Architecture of FineMAP and the corresponding auto-encoder network architecture: The orange part is encoder, the green part is decoder and the red part corresponds to the FineMap architecture}
  \label{fig:pipeline}
\end{figure}

For a CNN, the output of each convolution layer is the input of the next convolution layer,
\begin{align}
  f_{i+1} = conv(f_{i},\theta_i)
\end{align}
where, $f_{i}$ is the $i$th layer's output(feature map), $\theta_i$ is the parameters of the $i$th convolution layer and $conv$ is the operation of convolution. Among them, $f_{i}\in\mathbb{R}^{H,W,C}$, $H$ and $W$ is the height and width of the feature map and $C$ is the number of the feature maps.
This process continuously extract and transfer the features. Hence the feature map $f_{i}$ should be related to the target of the network. In ZSPAD, an auto-encoder is utilized to reconstruct the B-scans. Since B-scans are composed of invalid region and ROI, the $f_i$ should be related to the ROI.
Considering the correlation between the feature map $f$ of $F_{AE}$ and the ROI of $x$, we propose a fine-grained saliency map extraction method, namely FineMap.
Fig. \ref{fig:pipeline} shows the pipeline of FineMap. Because the output of decoder is more relevant to the reconstructed image, the feature maps are utilized for fine-grained map extraction. Specifically, we average the feature map along the channel dimension to obtain the saliency map of each layer,
\begin{align}
  MAP_i = \frac{1}{C}\sum_c^C resize(f_i^{H,W,c})
\end{align}
where $MAP_i$ is the saliency map of the $i$th layer, $resize$ is the resize function based on bilinear interpolation and $MAP_i \in \mathbb{R}^{H_x,W_x}$. $(H_x,W_x)$ is the size of the $x$. Then the fine-grained saliency map of the $x$ can be obtained by the process,
\begin{align}
  MAP_x = \frac{1}{I}\sum_i^I MAP_i
\end{align}
In this process, $MAP_x$ is the saliency map for $x$ and $I$ is the layer number of the decoder.
As can be seen, $MAP_x$ provides ROI based pixel-wise weight for both $x$ and $F^*_{AE}(x)$. The weight presents the importance of the corresponding pixel. Thus, the reconstruction error can be refined by,
\begin{align}
  \hat{R}^*_{AE}(x) = || F_{AE}(x,\hat{\theta}) \circ MAP_x  - x \circ MAP_x ||_2
\end{align}
where $\hat{R}^*_{AE}(x)$ is the refined reconstruction error and $\circ$ refers to the Hadamard product.
By using FineMap, the pixels in ROI have higher weight, while the pixels in invalid area have lower weight for the evaluation of the reconstruction error. Hence, the refined reconstruction error $\hat{R}^*_{AE}(x)$ is more precise and feasible for presentation attack detection.

Considering multiple B-scans (400 B-scans) obtained from one-scan, it is lack of robustness for presentation attack detection with single B-scan. Therefore we subsequently propose the specific confidence score for PAD using Gaussian Model based on multiple B-scans.
\subsection{Confidence Score for ZSPAD using Gaussian Model}
\label{sec:CSPAD}
Different from many shot task, ZSPAD can not utilize general classifiers(e.g., support vector machine and multilayer perceptron) for detection. On the other side, the data of PAD is out of the ordinary, i.e., multiple B-scans is derived at one-scan. Hence, in this section, we discussed the confidence score based on Gaussian Model. Specifically, Section \ref{sec:MBGM} presents the establishment of Gaussian Model using multiple B-scans. And we discussed the confidence score based on statistical feature of Gaussian Model in Section \ref{sec:CSSF} and the confidence score using probability density of Gaussian Model in Section \ref{sec:CSPD}.
\subsubsection{Multi B-scans based Gaussian Model}
\label{sec:MBGM}
In order to improve the robustness of the proposed ZSPAD model, we propose a Gaussian Model $G$. Considering multi B-scans obtained from only one-scan, we can define $G$ as,
\begin{align}
  G(\hat{R}^*_{AE}(x),m,s) = \mathbb{P}_G(x),\quad x\in S_{scan}
\end{align}
where $m$ is the mean value, $s$ is the standard deviation, $S_{scan}$ is the set of the obtained B-scans through one-scan and $\mathbb{P}_G$ is the probability density of the $\hat{R}^*_{AE}(x)$. Among them, both $m$ and $s$ can represent the statistical features of $\hat{R}^*_{AE}$ in $S_{scan}$.
When $S_{scan}$ is bonafide, $\hat{R}^*_{AE}$ is small and stable.
Thus, $m$ and $s$ is small. For PAIs, $\hat{R}^*_{AE}$ is larger and unstable. So, PAIs' $m$ and $s$ should be larger than the bonafides' $m$ and $s$. Hence, Gaussian Model is reasonable and feasible for ZSPAD.

To evaluate the $m$ and $s$ of the $\hat{R}^*_{AE}(x)$ in $S_{scan}$, we employ the maximum likelihood estimation(MLE) to optimize the $G$ and the objective function is given by,
\begin{align}
   \underset{m,s}{min} \quad \prod_{x\in S_{scan}}\mathbb{P}_G(x)
\end{align}
For each $S_{scan}$, we can get its analytical solution $m^*$ and $s^*$ through the above formula and we defined $m^*$ and $s^*$ as Mean Feature and STD Feature correspondingly. Hence, we obtain the optimal probability density function $\hat{\mathbb{P}}_G(x)$, i.e., Gaussian Model $\mathbb{P}_G(x)$ with $m^*$ and $s^*$.
Compared with $\hat{R}^*_{AE}(x)$, $\hat{\mathbb{P}}_G(x)$ consider multi B-scans for PAD and solve the problem caused by fluctuation of single image. The robustness of ZSPAD-Model is further improved.
\subsubsection{Confidence Score based on Statistical Feature}
\label{sec:CSSF}
In this section, we discussed the confidence score based on the statistical features $m^*$ and $s^*$ for presentation attack detection.Specifically, 4 confidence scores, S-Score, M-Score, S+M-Score and MS-Score are discussed in this section.

Through the Gaussian Model mentioned above, we can get $m^*_i$, $s^*_i \in S_{score}$. $S_{score}$ is the data set inlcuding bonafide without PAIs and $S_{score} \cap S_{model} = \emptyset$. $m^*_i$ and $s^*_i$ reflect the statistical feature (i.e., mean and standard deviation) corresponding to various bonafides. Hence it is utlilized for ZSPAD, we normalize them and define M-Score \& S-Score as,
\begin{align}
  Score_{s} &= \frac{|m_{test}-\overline{m}^*|}{m_{max}^*-\overline{m}^*}\\
  Score_{m} &= \frac{|s_{test}-\overline{s}^*|}{s_{max}^*-\overline{s}^*}
\end{align}
where $Score_{m}$ \& $Score_{s}$ correspond to M-Score \& S-Score, $m_{test}$ \& $s_{test}$ are the test samples' $m^*$ \& $s^*$ and $\overline{m}^*$ \& $\overline{s}^*$ are the averages of $m^*_i$, $s^*_i \in S_{score}$.
We define the distance between the farthest deviation value($m_{max}^*$ and $s_{max}^*$) and the center in $S_{score}$($\overline{m}^*$ and $\overline{s}^*$) as the denominator to normalize the $m^*_i$, $s^*_i \in S_{score}$.
The larger value of M-Score \& S-Score, the higher probability that subject is PA.
The thresholds(M-threshold and S-threshold) for M-Score and S-Score can be dynamically defined by the requirments for the risk.
This method effectively utilizes the data information with multi values of mean and standard deviation and both M-Score and S-Score have the same order of magnitude. Thus, we can define S+M-Score as,
\begin{align}
   Score_{S+M} = \frac{1}{2}(Score_{s} + Score_{m})
\end{align}
where $Score_{s+m}$ corresponds to S+M-Score. As can be seen, both feature $m^*$ and $s^*$ are considered by S+M-Score. However, simple addition may destroy their respective characteristics. And the weighted mean can not be utilized for this topic because of the Zero-PA setting.
Thus, we propose a novel strategy to achieve the joint judgment of these two features. We evaluate S-Score and M-Score separately, and build a voting model(MS-Score) for ZSPAD,
\begin{align}
   Score_{MS}\!=\!
\begin{cases}
\text{Bona}\!&\!{Score_{s}\!\leq\!S_{thres}, Score_{m}\!\leq\!M_{thres}}\\
\text{PAI}\!&\!{\text{others}}
\end{cases}
\end{align}
where Bona refers to the bonafide, $S_{thres}$ and $M_{thres}$ corresponds to S-threshold and M-threshold respectively.
When the S-Score of the subject is lower than the S-threshold and M-Score is lower than the M-threshold, the subject will be identified as bonafide.
Otherwise, the subject will be detected as PA, i.e., each score has the power of veto in this voting model.
MS-Score ensures the effective utilization of each feature and improves the performance of the ZSPAD-Model.
\subsubsection{Confidence Score based on Probability Density}
\label{sec:CSPD}
In this section, we present the confidence score based on probability density. Considering the relationship between $\hat{\mathbb{P}}_G(x)$ and the presentation attack detection, ZSPAD can be reduced to the evaluation of difference between $\hat{\mathbb{P}}^{score}_G(x)$ and $\hat{\mathbb{P}}^{test}_G(x)$. $\hat{\mathbb{P}}^{score}_G(x)$ is the Gaussian Model obtained from $S_{score}$ and $\hat{\mathbb{P}}^{test}_G(x)$ refers to the probability distribution of the test samples.
The stronger difference between the two probability distributions, the higher probability that subject is PA. We propose two methods to evaluate this difference degree. The first method utilizes the probability density directly, and it is defined as,
\begin{align}
  \text{PD-PostP} = \mathbb{E}(\hat{\mathbb{P}}^{score}_G(x_{test})), x_{test}\in S_{scan}\\
  \text{PD-PreP} = \hat{\mathbb{P}}^{score}_G(\mathbb{E}(x_{test})), x_{test}\in S_{scan}
\end{align}
where PD-PostP and PD-PreP is the confidence score for presentation attack detection and $x_{test}$ is the test B-scan. Both PD-PostP and PD-PreP utilizes the probability density of the $\hat{\mathbb{P}}^{score}_G(x)$. The larger PD-PostP and PD-PreP, the higher probability that subject is bonafide.
The difference between them is the evaluation order of the expectation for test sample.
\subsubsection{Confidence Score based on Divergence}
\label{sec:CSD}
The second method to evaluate the difference between $\hat{\mathbb{P}}^{score}_G(x)$ and $\hat{\mathbb{P}}^{test}_G(x)$ is based on the divergence. Specifically, Kullback-Leibler divergence (KL-divergence) and Intersection over Union (IoU) based divergence are utilized for confidence score. Since KL-divergence is asymmetric, we consider both case for KL-divergence and the confidence scores are defined as,
\begin{align}
\text{KL-Pre}  = KL(\hat{\mathbb{P}}^{score}_G||\hat{\mathbb{P}}^{test}_G)\\
\text{KL-Post} = KL(\hat{\mathbb{P}}^{test}_G||\hat{\mathbb{P}}^{score}_G)
\end{align}
where $KL$ refers to the KL-divergence, and KL-Pre \& KL-Post are the confidence scores based on KL-divergence. When KL-Pre and KL-Post is larger, the higher probability that test sample is PA.

For Gaussian Model, IoU based divergence function is also considered in this paper. The ratio of overlap region to total region of two Gaussian Models is usefule to present the difference between the two models. Therefore, we define an IOU-Score as,
\begin{align}
  \text{IOU-Score} = \frac{\hat{\mathbb{P}}^{score}_G \cap \hat{\mathbb{P}}^{test}_G}{\hat{\mathbb{P}}^{score}_G \cup \hat{\mathbb{P}}^{test}_G}
\end{align}
Since IOU-Score reflects the similarity between the two probability distribution, when the IOU score is large, the similarity between the two models is higher and the higher the probability that the test sample is a real sample.

The proposed confidence scores mentioned above have their own advantages and disadvantages. It is difficult to determine whether one of them is most suitable for ZSPAD.
Therefore, we carry out a series of experiments in the following Section to discuss the confidence scores.
\section{Experimental Results and Analysis}
\label{sec:ERA}
In this section, we present the results of the above proposed methods and the comparison with the learning based method \cite{liu2019high} \& feature based method \cite{chugh2019oct}.
In the Section \ref{sec:ID}, we give the implementation details for our proposed method. Section \ref{sec:EZSPAD} shows the ablation study and the effectiveness of the proposed confidence scores. Finally, in Section \ref{sec:Comparision}, we present the comparison with the existing methods.
\subsection{Data Set and Implementation Details}
\label{sec:ID}
\subsubsection{Zero-PA Setting and Data Set for ZSPAD}
PAD model based on Zero-Shot learning is quite different from traditional PAD model in terms of composition and setting of data set,
Hence we firstly provide a novelty difinition of the Zero-Shot setting, namely Zero-PA setting, and its data set for PAD problem.
\begin{figure}
  \centering
  \includegraphics[width=.99\textwidth]{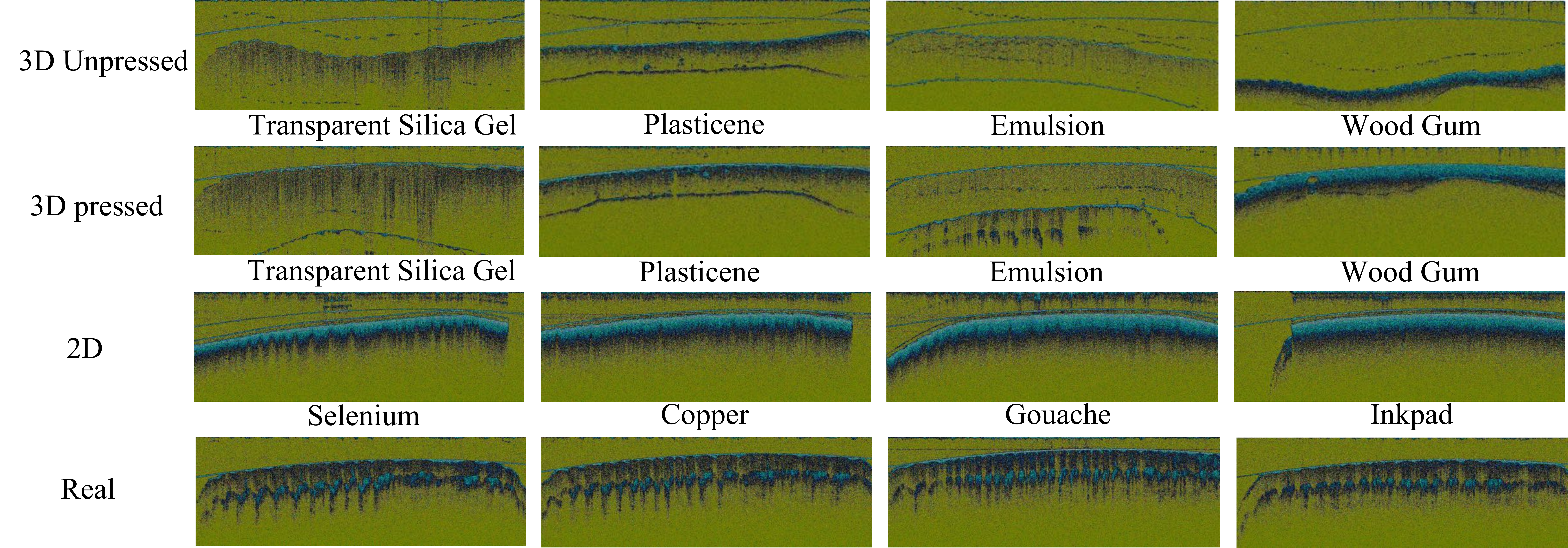}
  \caption{The various samples of B-scan, including bonafides and PAIs. The first row shows the B-scans scanned by 3D unpressed PAIs, the second row corresponds to the 3D pressed PAIs, the third row presents the 2D PAIs' B-scans and the fourth row give the samples from bonafides. The columns shows different instruments for the PAIs and various bonafide samples correspondingly.}
  \label{fig:samples}
\end{figure}

Let $S = \{ (x_i^{F}, y_i^{F}), (x_i^{R},y_i^{R})\}$ be the samples of PAD where $x_i^{F}$ is the B-scan of PA, $x_i^{R}$ is the B-scan of bonafide and $y_i$ is the corresponding class label(i.e., PA or bonafide).
For PAD, we partition $S$ into $S_{train}$ and $S_{test}$. Correspondingly, let $S_{train}= \{ (x_{train}^{R},y_{train}^{R}) \}$
and $S_{test} = \{ (x_{test}^{R},y_{test}^{R}), (x_{test}^{F},y_{test}^{F})\}$ be the partition of training and testing sets and $S_{test} \cap S_{train} = \emptyset$.
In this way, we partition the paired data into training and test set such that none of the PAs and bonafides from the test set occurs in the training set (same as traditional PAD settings).
On the other side, as shown in Fig. \ref{fig:zspad}(c), no PAIs are utilized in the process of ZSPAD model training and this is quite different from the traditional PAD settings.
Since the model has no access to any PAIs, the model needs to learn latent alignments between bonafides and PAs and generalize to perform well on the test data.

The model based on ZSL has a serious problems of bias, i.e., the probability distributions of the outputs with $S_{train}$ and $S_{test}$ are quite different.
Hence, it is difficult for ZSPAD model to define the confidence score, which is utilized to discriminate the class $y$ of the input $x$.
Specifically, it is infeasible to utilize the $S_{train}$ to define the confidence score like multi-shot learning method.
On the otherside, $S_{test}$ should not appear in the process of model establishment, including the determination of confidence score.
Thus, unlike other Zero-Shot settings, $S_{train}$ is further divided into Model Set $S_{model}$ and Score Set $S_{score}$.
The $S_{model}$ is utilized to train the ZSPAD model and $S_{score}$ is utilized to define the confidence score.
In order to guarantee the Zero-PA setting, $S_{score}$ is defined as a subset of $S_{train}$ rather than $S_{test}$, i.e., $S_{score}\cap S_{test} = \emptyset$.
And considering about the problem of bias in ZSL, we guarantee that $S_{score}$ and $S_{model}$ do not have any duplicate data, i.e. $S_{score} \cap S_{model} = \emptyset$.
The definition of $S_{score}$ requires the model to consider the gap between the $S_{train}$ and $S_{test}$. On the other side, a more objective confidence score with better generation ability can be determined using $S_{score}$.

In this paper, $S_{model}$ contains 41 finger samples derived from 41 bonafides to guarantee the diversity of the training data. $S_{score}$ consists of 16 finger samples scanned from only one bonafide(8 finegrs are scanned twice) to test the ZSPAD's generalization. $S_{test}$ is utilized to test the performance of PAD, which is composed of 176 finger samples scanned from 29 bonafides and 121 PA samples scanned from 121 PAIs. Fig. \ref{fig:samples} shows the samples of bonafides and PAIs. PAIs can be roughly divided into three categories, 3D unpressed, 3D pressed and 2D instruments.
3D PAIs are made by the 3D mold which is different from 2D's and the difference between 3D unpressed and pressed PAIs is whether it is pressed on the OCT acquisition device. In this paper, pressed and unpressed PAIs are considered as different PAIs. Thus, the 121 PAIs consist of 20 3D unpressed, 20 3D pressed and 81 2D PAIs.
For the size of data, each sample contains 400 B-scans, each B-scan consists of 1500 A-lines and the length of each A-line is 500 pixels. Therefore, each finger sample is a cube with size 400$\times$1500$\times$500.
\subsubsection{Implementation Details}
Our implementation is based on the public platform Tensorflow\cite{tensorflow2015-whitepaper}. We initialize the weights in each layer from a zero-mean Gaussian distribution with standard deviation 0.02. The Adam optimizer\cite{kingma2014adam} is utilized for our auto-encoder. Concretely, the learning rate of the Adam's is set to $5\times 10^{-5}$. $\beta_1$ and $\beta_2$ of Adam are set to 0.9 and 0.999 respectively.
Inspired by Maas et al.\cite{maas2013rectifier}, the leaky relu is utilized as the activation function for our approach.
Our workstation’s CPU is 2.8GHz, RAM is 32 GB and GPU is NVIDIA TITAN Xp.
\subsection{The Effectiveness of ZSPAD-Model}
\label{sec:EZSPAD}
\begin{figure}
  \centering
  \includegraphics[width=.99\textwidth]{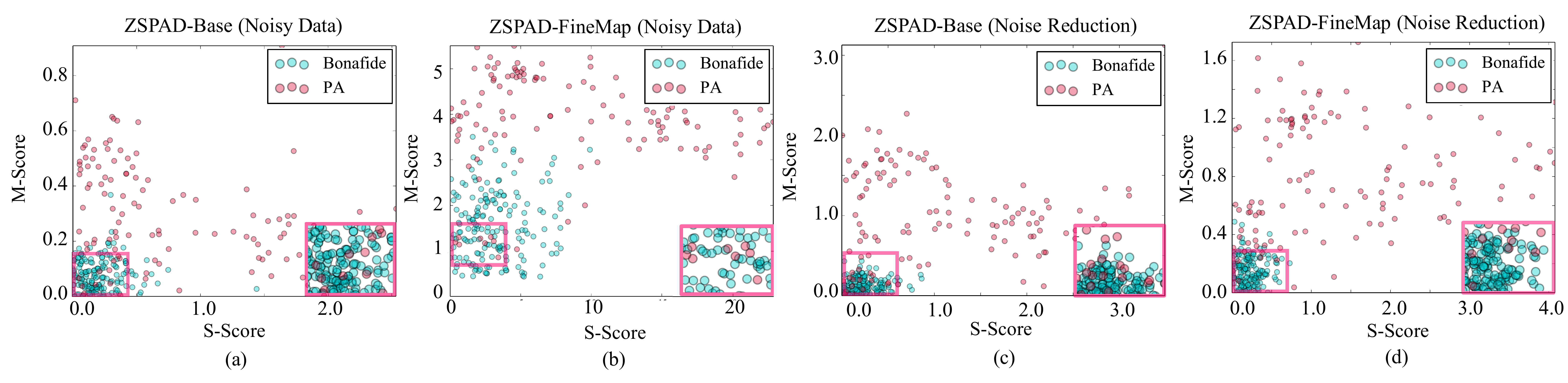}
  \caption{The ablation study of ZSPAD Model: the abscissa is S-Score, the ordinate is M-Score and each point corresponds to a sample(green points are the bonafides and red points are the PAs). ZSPAD-Base is our proposed baseline without ROI extraction and ZSPAD-FineMap is our proposed model using FineMap architecture.
  }
  \label{fig:score}
\end{figure}
Fig. \ref{fig:score} shows the distribution of the $S_{test}$ samples in the hyperspace of the M-Score and S-Score.
In the following part, as shown in Fig.\ref{fig:score}, we defined ZSPAD-Base as our baseline without refining and ZSPAD-FineMap as the proposed model using FineMap architecture.
Intuitively, in the confidence score space(see Fig.\ref{fig:score}), our proposed model aggregate the samples of the bonafides effectively. Taken as a whole, ZSPAD-Base and ZSPAD-FineMap have less outliers than ZSPAD-Patch, especially in the view of S-Score. Noted that in Fig. \ref{fig:score}(c)(d), the ZSPAD-FineMap has better generalization ability than ZSPAD-Base, which is reflected in the more uniform distribution of PAI points and the less number of the PA points are close to the bonafides.
On the other side, the influence of denosing for ZSPAD is important. Fig. \ref{fig:score}(a)(b) show the results with noise and Fig. \ref{fig:score}(c)(d) show the results without noise.
As can be seen, the noise reduction has a clustering effect on the bonafide points and promotes the performance of ZSPAD.
\begin{table}[!htbp]
\centering
\caption{The Ablation Study of ZSPAD-Model}
\resizebox{.98\textwidth}{9.2mm}{
\begin{tabular}{cccccccccccc} 
\toprule
~ &  \multicolumn{3}{c}{Statistical feature} &~& \multicolumn{3}{c}{Probability Density} &~&\multicolumn{3}{c}{Divergence} \\
\cline{2-4}
\cline{6-8}
\cline{10-12}
~ &Err. & TPR(\%)@FPR=10\% & TPR(\%)@FPR=5\% &~&Err. & TPR@FPR=10\% & TPR@FPR=5\% &~&Err. & TPR@FPR=10\% & TPR@FPR=5\% \\
  \midrule
ZSPAD-Base(noise)& 0.0741 & 92.05 & 75.00 &~&0.0744& 69.32 & 23.86 &~& 0.1616 & 71.59 & 57.39  \\
ZSPAD-FineMap(noise)& 0.0572 & 95.45 & 51.70 &~& 0.1010 & 26.70 & 6.25 &~& 0.3401 & 18.75 & 13.07\\
\midrule
ZSPAD-Base(denoise)& 0.0572 & 92.05 & 73.86 &~& 0.0640 & 96.02 & 55.11 &~& 0.0842 & 85.22 & 62.50 \\
ZSPAD-FineMap(denoise)& \textbf{0.0471} & \textbf{97.73} & \textbf{93.18} &~& 0.0572 & 97.16 & 63.64 &~& 0.1145 & 86.36 & 62.50 \\
  \bottomrule
\end{tabular}
}
\label{tab:Ablationall}
\end{table}
As shown in TABLE \ref{tab:Ablationall}, we make a quantitative analysis of our proposed model.
Three evaluation metrics are considered in this ablation study: Error(Err.), True Positive Rate(TPR)@False Positive Rate(FPR)=10\% and TPR@FPR=5\%. Among them, Err. reflects the best classification performance of the model.
The lower Err., the better the performance of the model is. TPR@FPR=10\% or 5\% represents the percentage of bonafides are able to access to the system when the accept rate of PAs $\leq$ 10\% or 5\%.
Statistical feature mentioned in TABLE \ref{tab:Ablationall} corresponds to the MS-Score, Probability Density refers to PD-PostP and Divergence is KL-Post. All of them empirically achieved the best result among their own kinds of confidence scores (i.e., statistical featture based, probability density based and divergence based confidence scores). From the TABLE \ref{tab:Ablationall} and Fig. \ref{fig:score}, we can draw the following conclusions:
(i) Confidence score plays a major role for the performance the ZSPAD-Model and statistical feature based confidence score (i.e., MS-Score) is the best confidence score for ZSPAD.
(ii) the denoising process is important for ZSPAD, which is remarkable in different score setting;
(iii) FineMap architecture is effective, which can further improve the precision of our proposed model.
(iv) ZSPAD-FineMap is the state of the art for Zero-Shot Presentation Attack, which achieves Err=0.0471, TPR = 97.73\%@FPR = 10\% and TPR=93.18\%@FPR = 5\% using MS-Score;

On the other side, experimental results show that the precision of the model is greatly affected by the confidence score. Hence, we discussed the relationship between confidence score and the performance of our proposed model in detail.
\begin{table*}[!htbp]
\centering
\caption{The Performance of ZSPAD-Model using different Confidence Score}
\resizebox{.76\textwidth}{20.2mm}{
\begin{tabular}{ccccc} 
\toprule
\multicolumn{2}{c}{Confidence Score}  &Err. & TPR(\%)@FPR=10\% & TPR(\%)@FPR=5\% \\
  \midrule
\multirow{4}{*}{Statistical Feature based} & S-Score & 0.1178 & 53.98 & 15.91 \\
&M-Score& 0.0539 & 96.59 & 81.25\\
&M $+$ S-Score&0.0673 & 92.05 & 86.36\\
&MS-Score&\textbf{0.0471} & \textbf{97.73} & \textbf{93.18}\\
\midrule
\multirow{2}{*}{Probability Density based}&PD-PostP& 0.0572 & 97.16 & 63.64\\
&PD-PreP& 0.0673 & 93.18 & 53.41  \\
\midrule
\multirow{3}{*}{Divergence based}&KL-Pre&0.2828 & 41.48 & 17.61 \\
&KL-Post& 0.1145 & 86.36 & 62.50\\
&IoU-Score&  0.0943 & 3.98 & 0\\
  \bottomrule
\end{tabular}
}
\label{tab:AblationConf}
\end{table*}
TABLE \ref{tab:AblationConf} shows the performance of our ZSPAD-Model(using FineMap) based on various case, including statistical feature, probability density and divergence based confidence score. As can be seen, different scores have a great influence on the results. Single S-Score can't achieve good results, but S-Score and M-Score are complementary, i.e., the performance of MS-Score is better than both of single scores. For probability density based confidence score, PD-PostP overperforms the PD-PreP. This indicates that the order of the expectation evaluation is important for the detection result. In the divergence based case, The confidence score can achieve the similar value to the above mentioned confidence scores in the term of Err.. But for more strict metrics, TPR@FPR=10\% and TPR@FPR=5\%, the performance of Divergence based confidence score is poorer than the others. In short, the preformance of both probability density and divergence based confidence scores are poorer than the MS-Score, which indicates MS-Score is more robust than the other confidence score and is empirically most suitable confidence score for ZSPAD.
\subsection{Comparison with Existing Methods}
\label{sec:Comparision}
In this section, we compare our model with existing methods. TABLE \ref{tab:ComparisionWithE} shows the result of the comparison. Since our best-performing model is given by FineMap and MS-Score mentioned in TABLE \ref{tab:Ablationall}, we use this model, namely ZSPAD-Model, by default in other parts of the paper unless noted. Learning based method \cite{chugh2019oct} depends on the training set, which includes both positive and negative samples, for parameter adjustment. The training set of the learning model consists of the PAIs from $S_{test}$ and the bonafides.
Therefore, the test set is composed of the unused samples in $S_{test}$.
The rate mentioned in TABLE \ref{tab:ComparisionWithE} is the percentage of the number of the utilized PAIs for training to the number of whole PAIs in $S_{score}$. The samples for training are selected randomly and the number of positive samples(i.e., bonafide) is same as the number of PAIs.
In order to ensure the fairness of the experiment, the test set of ZSPAD-Model, Feature Based Method and learning based method are the same.
\begin{table}[!htbp]
\centering
\caption{Comparison with Existing Methods}
\resizebox{.86\textwidth}{8.8mm}{
\begin{tabular}{cccccccc} 
\toprule
~ &  \multicolumn{3}{c}{@Rate=30\%} &~& \multicolumn{3}{c}{@Rate=50\%} \\
\cline{2-4}
\cline{6-8}
~ &Err. & TPR(\%)@FPR=10\% & TPR(\%)@FPR=5\% &~&Err. & TPR@FPR=10\% & TPR@FPR=5\%\\
  \midrule
Feature Based Method\cite{liu2019high}&0.1245 & 84.66 & 81.82 &~&0.1271 & 84.66 & 81.82 \\
Learning Based Method\cite{chugh2019oct}& 0.0667 & 90.55 & 90.55 &~&0.0472 & 95.18 & 94.81\\
ZSPAD-Model& \textbf{0.0311} & \textbf{100.00} & \textbf{97.73} &~& \textbf{0.0254} & \textbf{98.86} & \textbf{96.59} \\
  \bottomrule
\end{tabular}
}
\label{tab:ComparisionWithE}
\end{table}

As shown in TABLE \ref{tab:ComparisionWithE}, our proposed method overperforms the state of the art. In the case, where rate = 30\%, ZSPAD-Model achieves Err=0.0311, TPR = 100.00\%@FPR = 10\% and TPR=97.73\%@FPR = 5\% on the test set,
exceeding other methods Err = 0.0667/0.1245, TPR = 90.55/84.66\%@FPR = 10\% and TPR=90.55/81.82\%@FPR = 5\% by 0.0356/0.0934, 9.45/15.34\% and 7.18/15.91\% respectively. Our model is not only trained without any PAIs, but also surpasses existing methods significantly. This further proves the effectiveness of our model.

On the otherside, as shown in TABLE \ref{tab:ComparisionWithE}, the performance of the feature based method is stable but poor. This shows that the feature based method has a strong generalization, but the security is at weakly low level. And the performance of learning based method is improved with the increase of rate, which reflects its dependence on the data scale.
However, our proposed model does not have these issues. In order to further prove the effectiveness of our model, two different case with more details are provided in the following section. In Section \ref{sec:CFM}, we presents the detection result of feature based method and ZSPAD-Model for various PA. And we summarize the pros and cons of feature based method, focusing on its complementarity and low security. Section \ref{sec:CLM}) shows the data scale dependence of learning based method and the high-level generalization of ZSPAD-Model.
\begin{figure}
  \includegraphics[width=.99\textwidth]{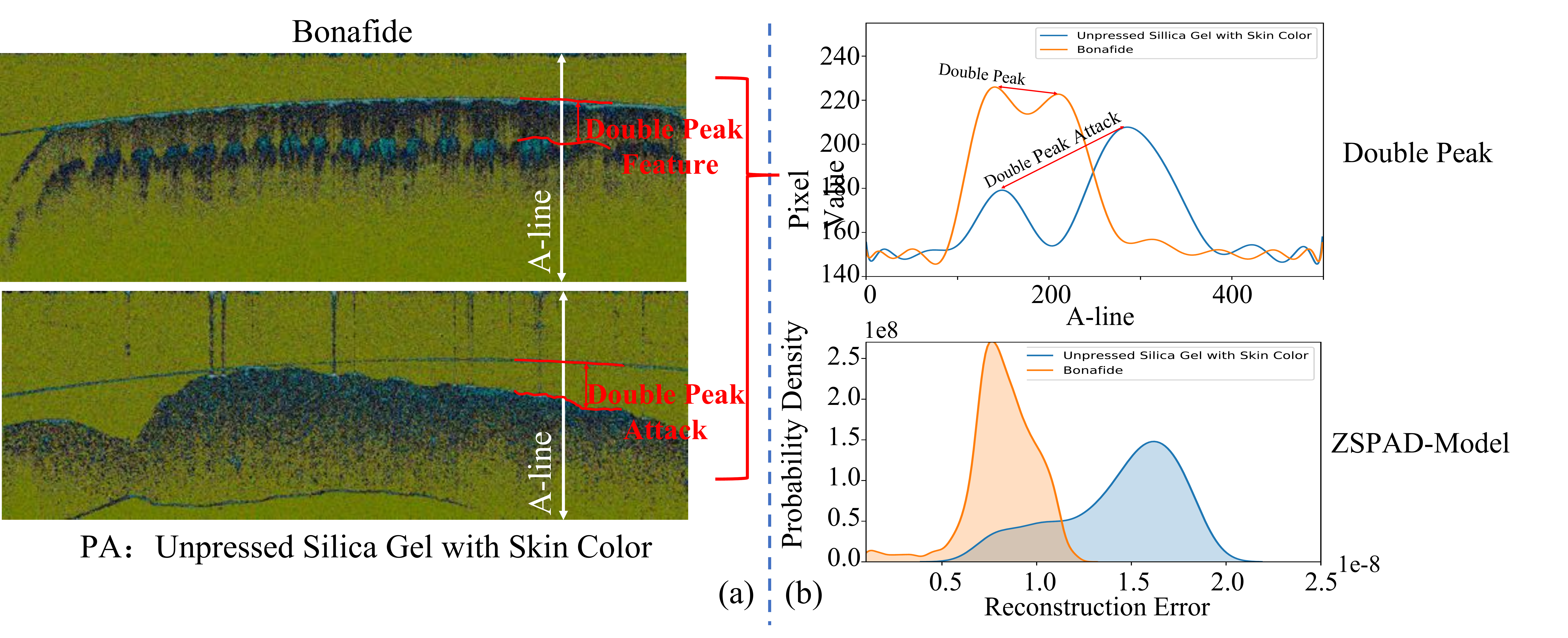}
  \caption{The special presentation attack for depth-double-peak feature based method.
            (a). The B-scans of bonafide and the presentation attack using unpressed silica gel.
            (b). First row: the real and deceptive double-depth-peak features represented in the A-line.
                 Second row: the gaussian distributions of the bonafide and unpressed silica gel using the proposed ZSPAD-Model.
           }
  \label{fig:attack}
\end{figure}
\subsubsection{Comparison with Feature based Method}
\label{sec:CFM}
Next we evaluate the feature based method \cite{liu2019high} in $S_{test}$. We discussed feature based method from two aspect, limitation and complementarity.

\textit{Limitation}: Liu et al. \cite{liu2019high} proposed method is based on feature. For this feature, we propose a special type PA, i.e., 3D unpressed PAIs. Fig. \ref{fig:attack} shows the special presentation attack for depth-double-peak feature based method. The strategy to attack this method is to utilize the unpressed silica gel to place it lightly on the acquisition device. The supported lense and the PAI form a deceptive depth-double-peak feature(see Fig. \ref{fig:attack}). Therefore, the method \cite{liu2019high} is difficult to identify this kind of PA. The security of feature based method is relatively low. Once the feature utilized for PAD is seen, the purpose of deception will be achieved from the kind of PAI or the acquisition method(e.g. the case we discussed in this section). But for our proposed method, it is hard to achieve PA using different types of attacks. In Fig. \ref{fig:attack}, the gaussian distributions of the bonafide and unpressed silica gel are quite different and can be detected easily.
\begin{table}[!htbp]
\centering
\caption{Comparison with Feature based Method for Various PAs}
\resizebox{.66\textwidth}{22.6mm}{
\begin{tabular}{cccc} 
\toprule
Case & Assessment & Feature based Method \cite{liu2019high} & ZSPAD-Model\\
  \midrule
  \multirow{3}{*}{2D PA} & Err. & 0.1042 & \textbf{0.0000} \\
                         & TPR(\%)@FPR=10\% & 0.8466 & \textbf{100.00}\\
                         & TPR(\%)@FPR=5\% & 0.8466 & \textbf{100.00} \\
  \midrule
  \multirow{3}{*}{3D-pressed PA} & Err. & 0.0872 & \textbf{0.0462} \\
                                 & TPR(\%)@FPR=10\% & 27.27 & \textbf{74.43}\\
                                 & TPR(\%)@FPR=5\% & 0.00 & \textbf{27.27} \\
  \midrule
  \multirow{3}{*}{3D-unpressed PA} & Err. & 0.0923 & \textbf{0.0410} \\
                                 & TPR(\%)@FPR=10\% & 32.95 & \textbf{81.82}\\
                                 & TPR(\%)@FPR=5\% & 18.18 & \textbf{52.84} \\
  \midrule
  \multirow{3}{*}{Average} & Err. & 0.1279 & \textbf{0.0471} \\
                                 & TPR(\%)@FPR=10\% & 84.66 & \textbf{97.73}\\
                                 & TPR(\%)@FPR=5\% & 53.98 & \textbf{93.18} \\
  \bottomrule
\end{tabular}
}
\label{tab:DoublePeak}
\end{table}

Meanwhile, we compare ZSPAD-Model and the feature based method on various cases. TABLE \ref{tab:DoublePeak} shows the results, and our proposed method overperforms the feature based method in every case remarkablely. Noted that, feature based method can achieve 100\% to detect the 2D PAIs, but in order to be compatible with 3D PAIs, we fine-tuning the hyper parameters(e.g. search scope) of the method.
\begin{table}[!htbp]
\centering
\caption{Complementarity between ZSPAD-Model and Feature based Method}
\resizebox{.66\textwidth}{8.6mm}{
\begin{tabular}{cccc} 
\toprule
~ & Err. & TPR(\%)@FPR=10\% & TPR(\%)@FPR=5\%\\
\midrule
Feature based Method\cite{liu2019high}&0.1279&84.66&53.98\\
ZSPAD-Model&0.0471&97.73&93.18\\
Fusion Method&\textbf{0.0438}&\textbf{100.00}&\textbf{96.02}\\
  \bottomrule
\end{tabular}
}
\label{tab:Fusion}
\end{table}

\textit{Complementarity}: On the other side, we discuss the complementarity between feature based method and ours. TABLE \ref{tab:Fusion} shows the performance of fusion method. The fusion method is on the basic of voting model. When ZSPAD-Model and feature based method recognize the input as bonafide at the same time, the final detection result is positive and other cases are detected as PA. As can be seen, fusion method overperforms any single method, and achieve 100\% in the TPR@FPR=10\%. This proves the high-level complementarity between feature based method and ours.
\subsubsection{Comparison with Learning based Method}
\label{sec:CLM}
\begin{table}[!htbp]
\centering
\caption{Comparison with Learning based Method with Various Train Rate of Data Set}
\resizebox{.68\textwidth}{25.6mm}{
\begin{tabular}{cccc} 
\toprule
~ & Err. & TPR(\%)@FPR=10\% & TPR(\%)@FPR=10\%\\
  \midrule
Learning Based Method@rate=5\% & 0.3645 & 4.96 & 1.33 \\
ZSPAD-Model@rate=5\%& \textbf{0.0481}& \textbf{96.59}  &  \textbf{91.48} \\
\midrule
Learning Based Method@rate=15\%  & 0.3407 & 29.06 & 26.25 \\
ZSPAD-Model@rate=15\%& \textbf{0.0386} & \textbf{98.86}  &  \textbf{93.75} \\
\midrule
Learning Based Method@rate=25\%  & 0.1457 & 73.27 & 62.59 \\
ZSPAD-Model@rate=25\%& \textbf{0.0358} & \textbf{98.86} & \textbf{96.59} \\
\midrule
Learning Based Method@rate=30\%  & 0.0667 & 90.55 & 90.55 \\
ZSPAD-Model@rate=30\%& \textbf{0.0311} & \textbf{100.00}  &  \textbf{97.73} \\
\midrule
Learning Based Method@rate=50\%  & 0.0472 & 95.18 & 94.81 \\
ZSPAD-Model@rate=50\%& \textbf{0.0254} & \textbf{98.86} & \textbf{96.59} \\
\midrule
Learning Based Method@rate=80\%  & \textbf{0.0043} & \textbf{99.68} & \textbf{99.63} \\
ZSPAD-Model@rate=80\%& 0.0323 & 97.73 & 91.48 \\
\bottomrule
\end{tabular}
}
\label{tab:Rate}
\end{table}

We now present the results of the learning based method \cite{chugh2019oct}. TABLE \ref{tab:Rate} shows the performance of the learning based method with different train rate. Specifically, in order to ensure the fairness of the experiment, the test set of ZSPAD-Model and learning based method are identical in each case.
As shown in TABLE \ref{tab:Rate}, the performance of learning based method is improved gradually with the increasing rate. And ZSPAD-Model is basically stable under various different test sets. This shows the method based on learning model has a strong dependence on the number of data and is lack of generalization. When rate = 50\%, learning based method achieves the similar result with ZSPAD-Model for the first time.
This proves that the performance of this unsupervised method(i.e., ZSPAD) is ideal, which can even surpass the supervised learning method in the case of small and medium data scale. There is no denying that with the increasing of training data, learning based method should overperform ours(rate = 80\%). However, it is difficult to collect PAIs in realistic application. So, in any case, our method is still practical and enlightening.
\begin{figure}
  \centering
  \includegraphics[width=.86\textwidth]{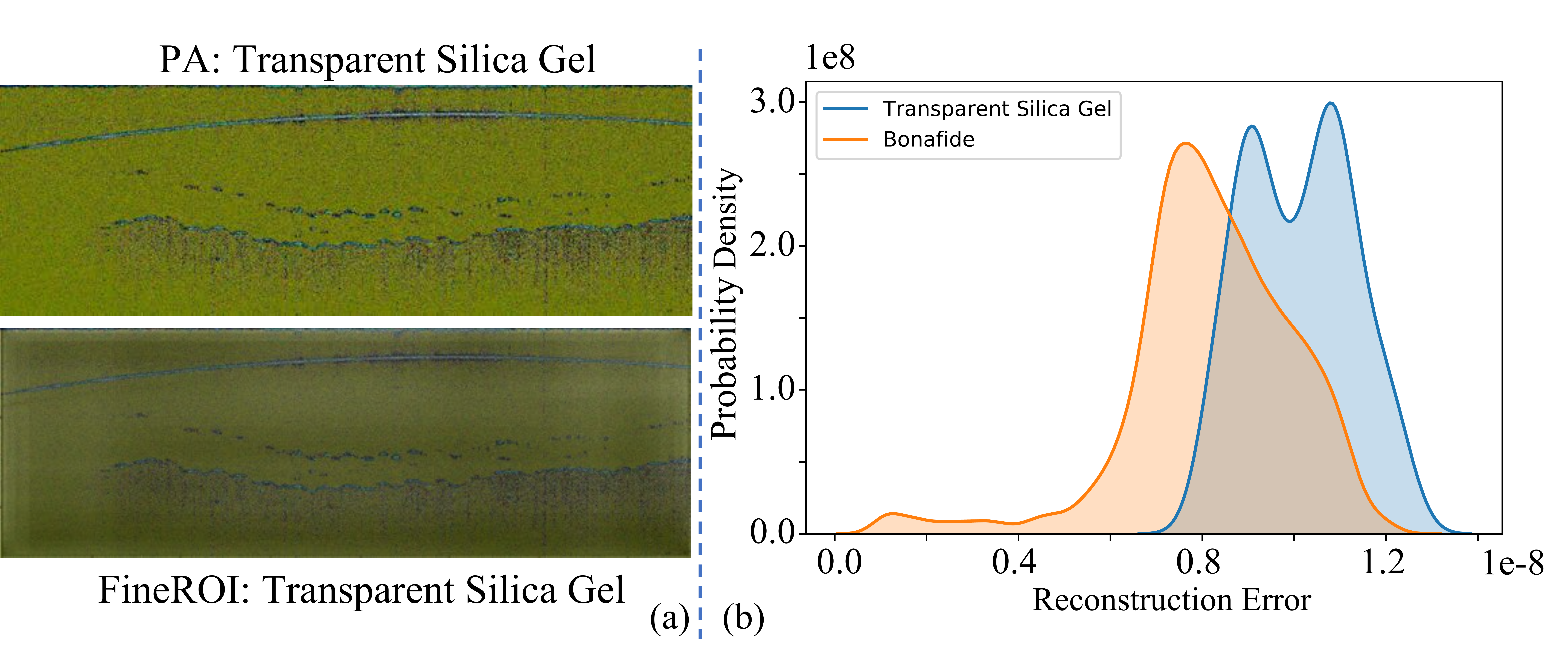}
  \caption{Failure case for ZSPAD-Model:
          (a) First row: the B-scan of PA made by transparent silica gel. Second row: the $ROI$ for the transparent silica gel using ZSPAD-Model
          (b) The gaussian distribution of the reconstruction error of the transparent silica gel on the basic of ZSPAD-Model.
           }
  \label{fig:chanllenge}
\end{figure}
\section{Conclusion \& Future Work}
\label{sec:Con}
This paper proposed a novel Zero-Shot Presentation Attack Detection framework to guarantee the generalization of the PAD model. By defining the Zero-PA setting, we urge the model to reduce the dependence on the presentation attack data. Specifically, an auto-encoder(AE) network is trained to reconstruct bonafide B-scans, and the trained model is utilized to evaluate the reconstruction error of PA and bonafide B-scan.
Since there is high-level noise and invalid areas in B-scan image disturbing the reconstruction error, we propose a Fine-grained Map architecture(FineMap) to provide a constraint for B-scan in order to refine the obtained reconstruction error.
Then we utilized the Gaussian Model to estimate the probability density of the reconstruction errors of a fingertips. Finally, three detection methods including, statistical feature, probability density and divergence based confidence scores are proposed for ZSPAD.
Experimental results showed that the ZSPAD-Model is the state of the art for ZSPAD, and the MS-Score is the best confidence score. Compared with existing methods,
our proposed method has stronger robustness, generalization ability and higher accuracy. Especially, in a small and medium data scale, the performance of ZSPAD-Model even exceeds the supervised learning method.

Although, our method has a good performance compared with the existing PAD method, TABLE \ref{tab:Ablationall} shows that the problem of PAD can not be solved completely by ZSPAD-Model. Hence, we analyzed the PAIs that failed to be detected and find that the failure causes are consistent. In this section, we take the PAI made by transparent silica gel as an example to discuss the limitation of our method. Fig. \ref{fig:chanllenge} shows this PAI and the result of our method. As can be seen, there is a small ROI in the B-scan of the transparent silica gel and this leads to the failure of the FineMap. Fig.\ref{fig:chanllenge}(a) presents the result of FineMap, and it can be seen that FineMap do not extract any valid information for ZSPAD-Model. The failure ROI extraction leads to the small reconstruction error, which leads to the similarity gaussian distribution between PA and bonafide(see Fig. \ref{fig:chanllenge}(b)). And this problem eventually influences the result of our method. Therefore, in our future work, we will further investigate how to reduce the dependence of our method on ROI extraction for more accurate ZSPAD.
\section{Acknowledgements}
The work is partially supported by the Shenzhen Fundamental Research fund
JCYJ2018
0305125822769,
the Education Department of Shaanxi Province (15JK1086).
In addition, we would like to thank Jiashu.Chen Haiming.Cao and Yong.Qi for their supports and contributions to this article.
\bibliographystyle{splncs04}
\bibliography{myreference}
\end{document}